# TL-SDD: A Transfer Learning-Based Method for Surface Defect Detection with Few Samples

Jiahui Cheng, Bin Guo*, Jiaqi Liu, Sicong Liu, Guangzhi Wu, Yueqi Sun, Zhiwen Yu
*Northwestern Polytechnical University*
Xi'an, China
cjhcjhcjh@mail.nwpu.edu.cn, {guob, jqliu, scliu}@nwpu.edu.cn, 13402806268@163.com, yqsun96@mail.nwpu.edu.cn, zhiwenyu@nwpu.edu.cn

*Abstract*—Surface defect detection plays an increasingly important role in manufacturing industry to guarantee the product quality. Many deep learning methods have been widely used in surface defect detection tasks, and have been proven to perform well in defects classification and location. However, deep learning-based detection methods often require plenty of data for training, which fail to apply to the real industrial scenarios since the distribution of defect categories is often imbalanced. In other words, common defect classes have many samples but rare defect classes have extremely few samples, and it is difficult for these methods to well detect rare defect classes. To solve the imbalanced distribution problem, in this paper we propose TL-SDD: a novel Transfer Learning-based method for Surface Defect Detection. First, we adopt a two-phase training scheme to transfer the knowledge from common defect classes to rare defect classes. Second, we propose a novel Metric-based Surface Defect Detection (M-SDD) model. We design three modules for this model: (1) feature extraction module: containing feature fusion which combines high-level semantic information with low-level structural information. (2) feature reweighting module: transforming examples to a reweighting vector that indicates the importance of features. (3) distance metric module: learning a metric space in which defects are classified by computing distances to representations of each category. Finally, we validate the performance of our proposed method on a real dataset including surface defects of aluminum profiles. Compared to the baseline methods, the performance of our proposed method has improved by up to 11.98% for rare defect classes.

I. INTRODUCTION

Surface defect detection plays an increasingly important role manufacturing industry to guarantee the product quality. The automated product surface defect detection methods can timely find and control the adverse effects of these defects on the aesthetics and performance of the product. Generally, there are four lines of surface defect detection methods: (1) eddy current detection, (2) infrared detection, (3) magnetic flux leakage detection and (4) computer vision detection. The former three methods are far from satisfactory in terms of a wide variety of defect detection effect due to their insufficient application. Computer vision detection methods gradually take the place of the artificial detection methods and realize the automated surface defect detection, which also benefits a wide range of intelligent manufacturing domains, including machinery manufacturing [1], aerospace [2] and other fields.

Surface defect detection consists of two components: (1) Defect classification – to detect the types of defects exist in the image. Traditional surface defect classification methods often use conventional image processing algorithms or artificially designed features based classifiers. Image processing methods generally include image preprocessing [3] and image segmentation [4, 5]. Manual image features usually include texture features [6, 7], color features, shape features, etc. The features are extracted from the image and then be input into the classifier for defect classification. (2) Defect location - not only to detect the types of defects exist in the image, but also to locate these defects. Many deep learning-based methods are widely used in surface defect detection tasks, which can not only improve the performance of defect classification, but also realize defect location [8]. In some methods, a candidate box is generated and then categorized for the defect location. In others, a convolutional neural network (CNN) is leveraged to directly predict the category and location of defects.

Though deep learning-based methods perform well in some cases, they fail to work in real industrial scenarios since the distribution of defect categories is often imbalanced, that is, common defect classes have many samples but rare defect classes have extremely few samples, as shown in Figure 1. It is difficult for these methods to well detect rare defect classes. In real industrial environment, due to the limitation of the production environment, there widely exist defect categories with a large number of samples and defect categories with extremely few samples. If we train all the data together as usual, we can obtain a detection model with good overall performance but poor performance for the rare defect classes with extremely few samples. In addition, the feature map obtained through general feature extraction will lose structural information due to convolution and pooling operations. Since defects vary greatly, the same feature extraction operation is not effective for different defects, especially for small range of defect categories. In real production scenarios, surface defect detection still faces the following challenges:

- The distribution of defect categories is extremely imbalanced, which means there are plenty of common defect samples but extremely few rare defect samples.

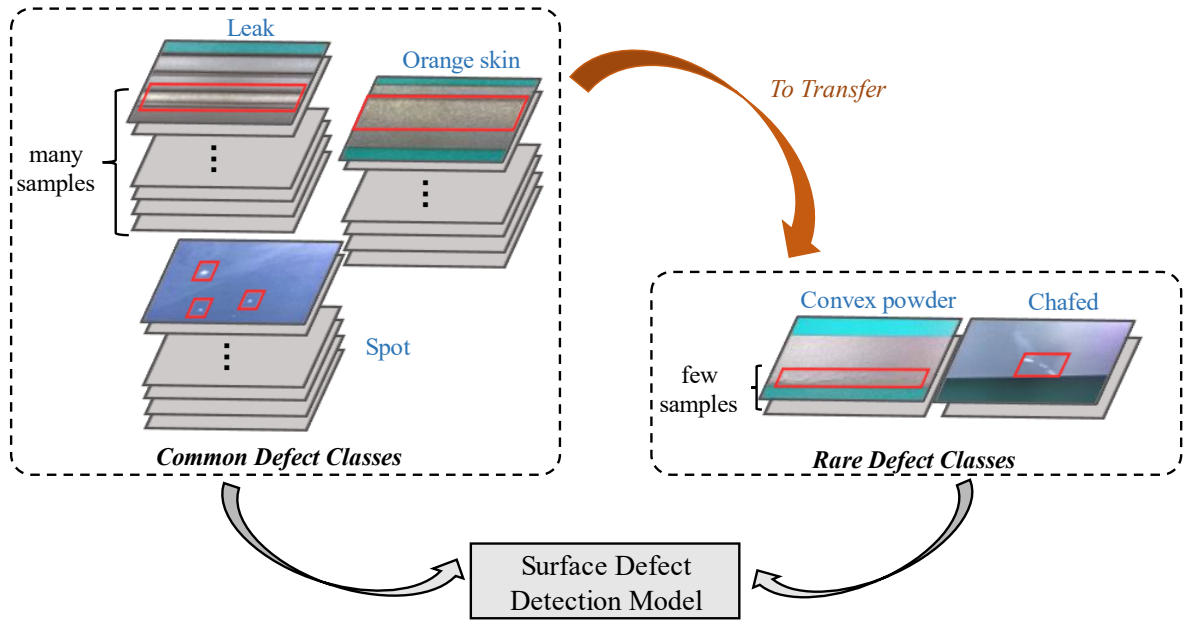

Fig. 1.Imbalanced distribution of defect categories: common defect classes with plenty of samples and rare defect classes with extremely few samples.

When these imbalanced samples are carried out in joint training, it is disadvantageous for rare defect classes.

- It is difficult to detect small range of defect categories. Because of convolution and pooling operations, the feature map from the general feature extractor loses structural information.

To address these issues mentioned above, in this paper we propose a Transfer Learning-based method for Surface Defect Detection (TL-SDD) which contains a two-phase transfer learning scheme and a novel Metric-based Surface Defect Detection (M-SDD) model. The two-phase transfer learning scheme transfer the knowledge from common defect classes to rare defect classes. The first phase is to obtain a pre-trained defect detector with common defect samples, then finetune it with few both rare defect and common defect samples in the second phase. Feature fusion is adopted in feature extraction module to prevent the loss of structural information. To leverage plenty of common defect classes and quickly adapt to rare defect classes, we design a distance metric module to categorize the defects instead of fully connected network. Besides, we add a feature reweighting module trained in parallel with feature extraction module. This module transforms examples to a reweighting vector that indicates the importance of features. Overall, our contributions can be briefly summarized as follows:

- We design a two-phase transfer learning scheme to ensure the generalization performance of the model for the rare defect classes with extremely few samples.
- We design a novel surface detection model, which contains feature extraction module, feature reweighting module and distance metric module.
- We conduct experiments on a real dataset including surface defects of aluminum profiles. The detection performance of our method has improved by up to 11.98% compared to the baseline methods.

The remaining parts of this article are depicted as follows. The related work about this work is presented in Section II. In Section III, a novel surface defect detection method is provided. The experimental results on different surface defect detection situations are given in Section IV. Finally, a brief summary is presented in Section V.

## II. RELATED WORK

### A. Surface defect detection

Sliding window. Generally, redundant sliding is performed on the original image through a smaller-sized window, and the image in the sliding window is input into the classification network for defect recognition. Finally, all the sliding windows are connected to obtain the result of rough positioning of the defect. Cha [9] et al. used the sliding window-based CNN classification network to realize the crack surface defect location. The limitation of this kind of method is that the sliding traversal speed is slow, the sliding window size needs to be selected accurately, and only a coarser-grained positioning effect can be obtained.

Heatmap. A heatmap is an image that reflects the importance of each area in the image. In the field of computer vision, CAM (Class Activation Mapping) [10] and Grad-CAM [11] methods are often used to obtain heatmaps. Lin [12] et al. used CAM to obtain the heatmap, and used the Otsu binarization method to segment the heatmap to realize the location of scratches or line defects in the LED lamp image. Zhou [13] et al. used the grad-CAM method to obtain the heat map, and also used the Otsu algorithm to segment to obtain the accurate area of the surface of defect location depends on the network classification performance.

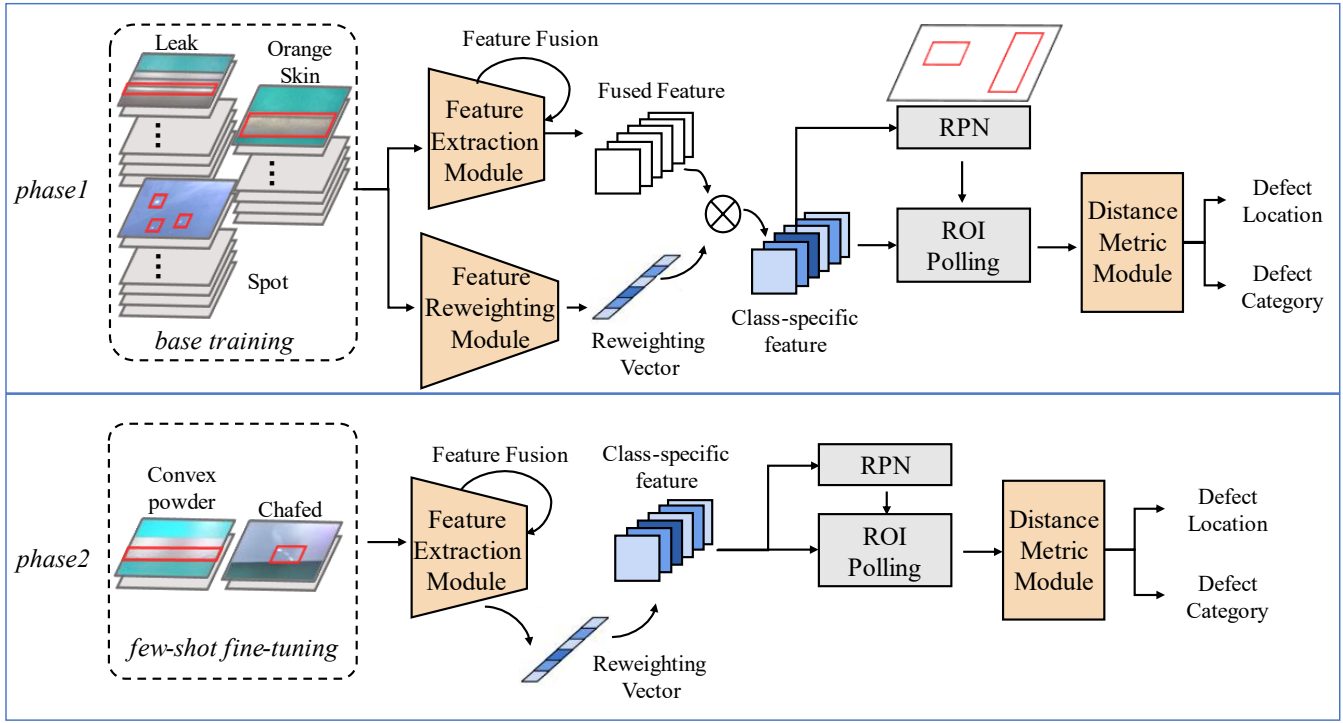

Fig. 2.TL-SDD Framework. Our proposed model leverages fully labeled common defect classes and quickly adapts to rare defect classes, using a feature extraction module , a feature reweighting module and a distance metric model within a two-phase transfer learning scheme.

Object detection network. Object detection is one of the most basic tasks in the field of computer vision, and it is also the closest task to defect detection in the traditional sense. At present, object detection methods are based on deep learning emerge in an endless stream. Generally speaking, the defect detection network based on deep learning can be divided into two categories: (1) two-stage network represented by Faster R-CNN [14]. (2) one-stage network represented by SSD [15] or YOLO [16]. The two-stage method has higher detection accuracy. This work leverages Faster R-CNN as the backbone to improve and realize surface defect detection.

### B. Surface defect detection with few samples

Data Augmentation. The most commonly used defect image amplification method leverages multiple image processing operations such as mirroring, rotation, translation, and distortion on the original defect samples to obtain more samples. Huang [17] et al. used the above method to amplify the defect data and apply it to the magnetic tile defect detection. Another common method is data synthesis. Tao [18] et al. used a segmentation network to segment the defective insulator from the natural background, and then superimposed it on the normal sample through image fusion. In this paper, we use some data augmentation methods to expand the dataset.

Weakly supervised learning. Usually the method based on weakly supervision refers to the use of image-level category annotation to obtain the detection effect of sub-location level. Marino [19] et al. used a weakly supervised learning method based on PRM (Peak Response Maps) to classify, locate and segment potato surface defects. Niu [20] et al. proposed a weakly supervised learning defect detection method based on GAN. Through CycleGAN [21], the input test image is converted to its corresponding defect-free image, and the difference between the input image and the generated defect-free image is compared to realize the surface defect detection. This type of method still requires a large number of labeled images, and is not suitable for situations with extremely few data.

Semi-supervised learning. Semi-supervised learning usually uses a large amount of unlabeled data and a small part of labeled data for the training of surface defect detection models. Di [22] et al. proposed a semi-supervised GAN [23] network-based method to classify steel surface defects. In the designed CAE-GAN defect detection network, a CAE-based encoder is used and fed into the softmax layer to form a discrimination device. He [24] et al. proposed a multi-trained semi-supervised learning method applied to the classification of steel surface defects. This method uses cDCGAN [25] to generate a large number of unlabeled samples. Gao [1] et al. proposed a semi-supervised learning method using CNN and improved the performance by using pseudo labels. At present, semi-supervised methods are mostly used to solve defect classification tasks, and have not been widely used in defect location.

Meta-learning. The main goal of meta-learning is to learn prior knowledge and then be leveraged to quickly adapt to a new task. According to the process of learning parameters in the adaptive process, meta-learning methods can be divided into three types: (1) Optimization-based meta-learning: Finn [26] et al. proposed a model-independent meta-learning method MAML, which can learn a relatively good initialization parameter, so that the model can be quickly fine-tuned to have a relatively good effect after accepting a new task. (2) Model-based meta-learning: Santoro [27] et al. proposed a neural network with memory enhancement called MANN, which uses external memory space to explicitly record some information and combines it with the long-term memory capabilities of the neural network itself to realize few samples learning tasks. (3) Metric-based meta-learning: Snell [28] et al. proposed prototype networks, which learn a metric space in which classification can be completed by calculating the distance to the prototype representation of each class. This paper draws on the metric-based meta-learning method to realize the detection of surface defects with imbalanced categories and few samples.

### III. APPROACH

In this paper, we name the defect categories with plenty of annotated data as common defect classes and those with extremely few annotated samples as rare defect classes. We aim to obtain a surface defect detection model which can learn to well detect both common defect classes and rare defect classes by transferring knowledge from common defect classes to rare ones.

Our proposed surface defect detection model introduces a feature extraction module $E$, a feature reweighting module $P$, and a distance metric module $M$ into a two-stage detection framework Faster R-CNN [14]. And we propose a two-phase transfer learning scheme to ensure the generalization of the model for the rare defect classes. This setting is in good agreement with the reality -- one might expect to deploy a pre-trained defect detector for rare defect classes with extremely few labeled samples. Our framework is shown as Figure 2.

#### A. M-SDD model

*1) Feature extraction module*

We use ResNet-101 [29] to implement the backbone of feature extraction module $E$. In this module, we use the Feature Pyramid Networks [30] (FPN) for feature fusion. As shown in Figure 3, The left to right path is the feedforward calculation of the backbone CNN, and the feature hierarchy is composed of multi-scale feature maps is calculated with a scaling step of 2. The right to left path produces higher-resolution features by upsampling of more coarse-grained space. Because all levels of the pyramid use shared classifiers/regressors like traditional feature image pyramids, we use the same feature dimensions in all feature mappings. To keep the same number of channels at each layer, use 1×1 convolution to change channel dimensions. In order to be fused with high-resolution features, we increase the spatial resolution of low-resolution features by a factor of 2. In order to reduce the aliasing effect of upsampling, we append 3×3 convolution to each merged map to generate the final fused feature.

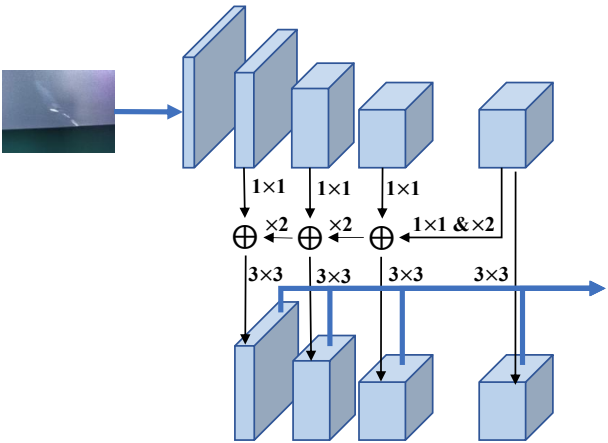

Fig. 3. Feature extraction module. The left to right path at the top is the feedforward calculation of the backbone CNN. The right to left path in the middle is feature fusion. 1×1 convolution is to change channel dimensions. ×2 means upsampling. 3×3 convolution is to reduce the aliasing effect of upsampling.

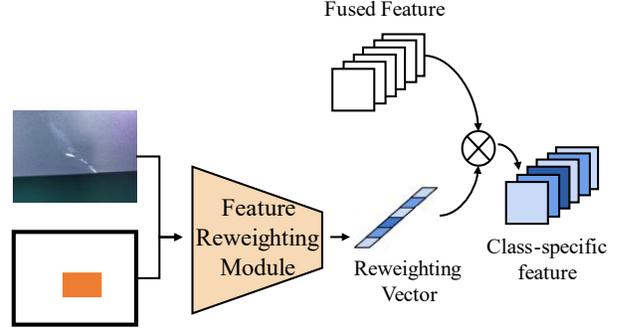

Fig. 4. Feature reweighting module. This module takes the defect images and their annotations as input, learns to embed these information into the reweighting vector, and adjusts the contribution of fused feature.

*2) Feature reweighting module*

For the feature reweighting module $P$, we design a lightweight CNN, which can not only improve efficiency, but also simplify learning. The schematic diagram of this module is shown as Figure 4. This module takes the defect images and their annotations as input, learns to embed these information into the reweighting vector, and adjusts the contribution of the fused feature, so as to be used by subsequent modules.

Formally, let $F \in \mathbb{R}^{w \times h \times m}$ denotes fused feature, which is generated by $E: F = E(x)$, where $x$ denotes the input defect images. The fused feature has $m$ feature maps. $x_i$ and $a_i$ denotes defect images and their associated bounding box annotations respectively, for class $i, i = 1, \dots, N$. The reweighting module takes $(x_i, a_i)$ as input, embeds it into a reweighting vector $w_i \in \mathbb{R}^m$ with $w_i = P(x_i, a_i)$. The vector is responsible for adjusting the weight of fused features and highlighting more important features in class $i$. After getting the vector $w_i$, our model applies it to obtain the class-specific feature $F_i$ for class $i$ by:

$$F_i = F \otimes w_i \qquad (1)$$

where $\otimes$ denotes 1×1 depth-wise convolution.

After obtaining the class-specific feature $F_i$, RPN network and ROI pooling will generate candidate Region of Interest (ROI) sets of the same size. Please refer to Faster R-CNN[14] for details.

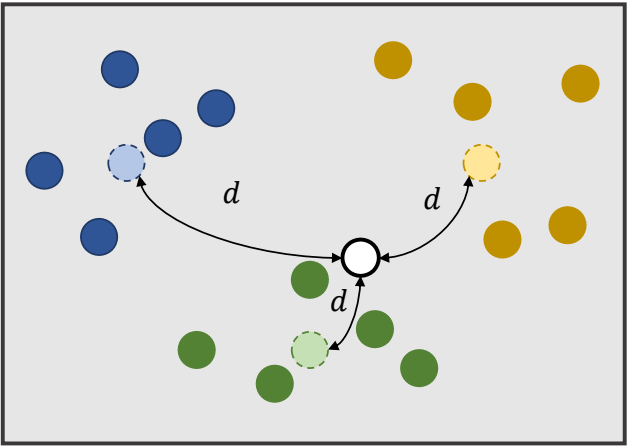

Fig. 5. Distance metric module. This module learns a metric space in which the representation of each defect category is generated and defects are classified by computing distances to representations of each category.

*3) Distance metric module*

In distance metric module $M$, we generate a representation for each defect category, which is classified by using softmax function to calculate the probability that the sample belongs to each. The schematic diagram of this module is shown as Figure 5. In this way, only a few parameters can be used for classification, so as to avoid over-fitting of rare defect classes with few training samples when using the fully connected network.

We have dataset $D = D_1 \cup ... \cup D_N$, class $i, i = 1, ..., N$. $D_i = \{(x_1^i, y_1^i), ..., (x_n^i, y_n^i)\}$ denotes a subset of the dataset $D$ belonging to class $i$. We define a function $f$, consisting of feature extractor, feature reweighting module, RPN and ROI pooling. Distance metric module compute a representation $c_i$ for each class $i$ by:

$$c_i = \frac{1}{|D_i|}\sum_{(x,y)\in D_i} f(x) \quad (2)$$

After the representation of each class is obtained, the probability of each sample corresponding to each class is calculated by:

$$P(y = i|x) = \frac{\exp(-d(f(x),c_i))}{\sum_{i'} \exp(-d(f(x),c_{i'}))} \quad (3)$$

where distance function $d$ means squared Euclidean distance.

Finally, we choose loss function for the whole model as:

$$L = L_{loc} + L_{cla} \quad (4)$$

where $L_{loc}$ is similar to loss function in RPN and

$$L_{cla} = -\log P(y = i|x) \quad (5)$$

*B. Two-phase transfer learning scheme*

To ensure the generalization performance of the model, we proposed a two-phase transfer learning scheme which is different from the traditional model. We reorganize the annotated training images in the common defect classes into several few-shot defect detection learning tasks $T_k$. Each task $T_k = S_k \cup Q_k$ contains a support set $S_k$ (consisting of $N \times s$ support images, which means each class has $s$ images, class $i = 1, ..., N$) and a query set $Q_k$ (offering query images with annotations for performance evaluation).

The whole learning process consists of two phases. (1) The first phase is the base training phase. In this phase, we jointly train the feature extractor, feature reweighting module, RPN and distance metric module with abundant few-shot tasks made up of common defect classes samples. In each task, we randomly sample $s(s=5)$ labeled images for each class in support set and each task and 2 labeled images for each class in query set of each task. (2) The second phase is few-shot fine-tuning. In this phase, we train the model on both common defect classes and rare defect classes. To keep balance between common defect classes and rare defect classes, we only include $s(s=5)$ labeled images for both common defect classes and rare defect classes in support set and 2 labeled images for each class in query set. We freeze the feature extractor in the training procedure and finetune the rest module.

IV. EXPERIMENTS

In this section, we evaluate our model in a real dataset including surface defects of aluminum profiles. We use Faster R-CNN [14] as the base detector and choose ResNet-101 [29] as backbone.

*A. Experimental Setup*

*1) Dataset*

In this paper, we use a real dataset including surface defects of aluminum profiles to evaluate our model. This dataset is provided by the Guangdong Industry Intelligent Manufacturing Big Data Innovation Competition. We choose 4 common defect classes (Leak, Pit, Orange skin, Spot) and 2 rare defect classes (Convex powder, Chafed). This paper focuses on a single defect on the surface of aluminum, where a single defect refers to only a certain kind of defect on a single image. Because of the small number of images in the original dataset, we make data augmented by horizontal mirroring, vertical mirroring and 180° inversion of the original image. The schematic diagram of data augmentation is shown as Figure 6. Finally, each category in common defect classes contains at least 1000 images and each category in rare defect classes contains 64 images. The details of the dataset are shown in Table 1.

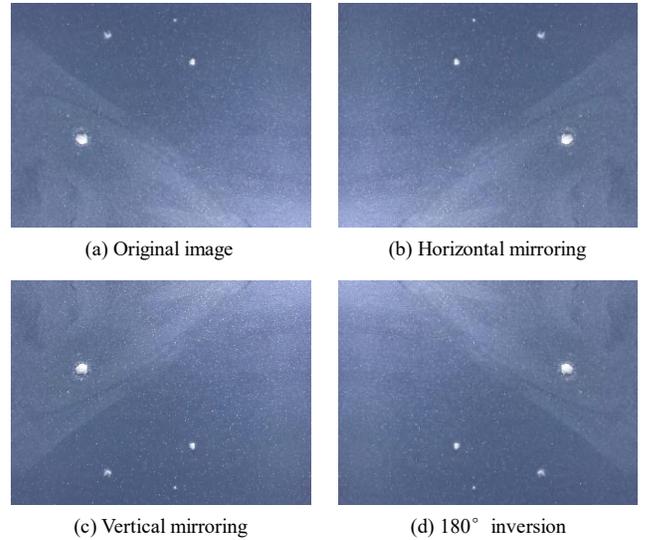

(a) Original image  (b) Horizontal mirroring

(c) Vertical mirroring  (d) 180° inversion

Fig. 6. Data augmentation. (a) is the original image, (b) is horizontal mirroring of the original image, (c) is vertical mirroring of the original image and (d) is 180° inversion of the original image.

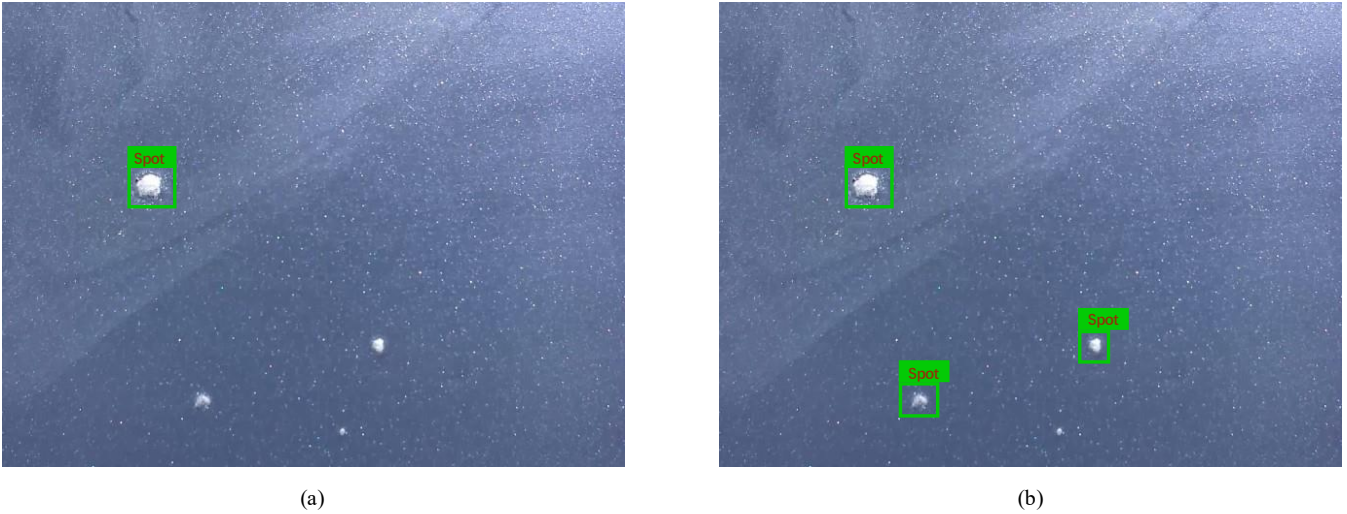

(a)                                      (b)

Fig. 7. Comparison of the two methods (*Faster R-CNN-joint* (*FR-joint*) and *Faster R-CNN+ff-joint* (*FR+ff-joint*) ) about the detection results of spot defects. (a) is the detection results of spot defects *with FR-joint*. (b) is the detection results of spot defects with *FR+ff-joint*.

TABLE I.       THE DETAILS OF SURFACE DEFECTS DETECTION OF ALUMINUM PROFILES DATASET

| Common defect | | | | Rare defect | |
|---|---|---|---|---|---|
| *Leak* | *Pit* | *Spot* | *Orange skin* | *Convex powder* | *Chafed* |
| 1076 | 1228 | 1044 | 1092 | 64 | 64 |

*2) Evaluation metrics*

In this paper, we choose Average Precision (AP) for each category as evaluation indicator of our model. The indicators are identified as:

$$Precision = TP/(TP + FP) \qquad (6)$$

$$Recall = TP/(TP + FN) \qquad (7)$$

$$AP = (Precision + Recall)/2 \qquad (8)$$

where TP (True Positives) denotes the number of positive samples that are correctly identified as positive samples, FP (False positives) denotes the number of negative samples that are misidentified as positive samples, and FN (False negatives) demotes the number of positive samples that were misidentified as negative samples.

*3) Baselines*

We compare our model with four baselines. *Faster R-CNN-joint* (*FR-joint*) is the first baseline which train the detector with images from the common and rare defect classes together on Faster R-CNN. The second one is *Faster R-CNN+ff-joint* (*FR+ff-joint*) which adds feature fusion in feature extraction module and trains the detector on images from the common and rare defect classes together. The third one is two-phase Transfer Learning scheme adds feature fusion (*TL-ff*) in feature extraction module which trains Faster R-CNN with feature fusion on samples of common defects and then finetunes it on all defects. The fourth one (*TL-ff+fr*) adds a feature reweighting module on the basis of the third one.

All the above baselines use Fast R-CNN [31] to classify defects, while our method uses the distance metric module. Stochastic gradient descent (SGD) with momentum is used for model training with a base learning rate of 0.0001.

TABLE II.       EXPERIMENTAL RESULTS ON THREE METHODS

| AP(%) | Common defect | | | | Rare defect | |
|---|---|---|---|---|---|---|
| | *Leak* | *Pit* | *Spot* | *Orange skin* | *Convex powder* | *Chafed* |
| *FR-joint* | 74.83 | 69.87 | 54.69 | 75.53 | 46.01 | 49.08 |
| *FR+ff-joint* | 76.07 | 72.98 | 62.68 | 76.91 | 46.84 | 50.82 |
| *TL-ff* | 76.11 | 73.02 | 61.97 | 76.85 | 50.23 | 52.83 |
| *TL-ff+fr* | 76.39 | 73.06 | 66.73 | 77.69 | 53.58 | 55.32 |
| **TL-SDD** | **76.96** | **73.09** | **68.17** | **78.02** | **59.78** | **61.84** |

*B. Experimental results*

We present our main results in Table 2. We note that our model significantly outperforms the baselines. Compared with the first baseline (*FR-joint*), the second baseline (*FR+ff-joint*) adds feature fusion in feature extractor. *FR+ff-joint* brings 2.88% improvement compared to *FR-joint* because of the feature fusion. Especially for spot defects, the detection performance is improved by 2% after feature fusion. This is because the scope of spot defects is small, and feature fusion combines high-level semantic information and low-level structural information, which greatly improves the detection effect of spot defects. Comparison of the two methods about the detection results of spot defects is shown in Figure 7. With the addition of feature fusion, the model can detect the dirty spot defects more effectively.

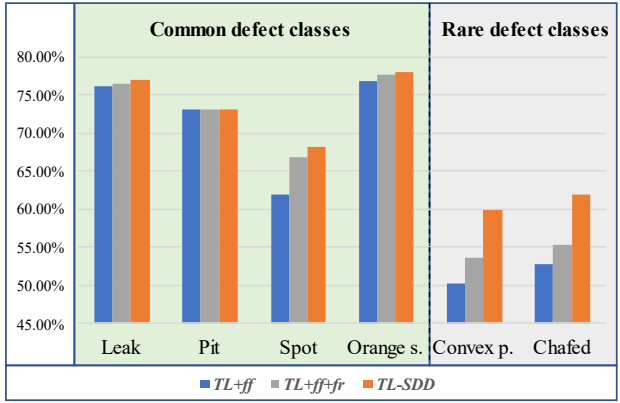

Fig. 8. The performance of *TL-ff*, *TL-ff+fr* and *TL-SDD*. The first two lines at the top belong to rare defect classes. The rest of the lines belong to common defect classes. The orange column shows the performance of our method *TL-SDD*. The grey one and blue one shows the performance of *TL-ff+fr* and *TL-ff* respectively.

Compared with the first two baselines, the last three methods in Table 2 trains the detector with two-phase learning scheme. The performance of our method (*TL-SDD*) was improved by up to 11.98% for rare defect classes. This reflects the necessity of the two training phases employed in our model: it is better to first train a good model on common defect classes and then fine-tune with few-shot data, otherwise joint training will let the detector bias towards common defect classes and learn nearly nothing about rare defect classes.

Figure 8 shows the performance of the last three methods in Table 2 including *TL-ff*, *TL-ff+fr* and *TL-SDD* (our method). Compared with *TL-ff*, *TL-ff+fr* adds a feature reweighting module which improve the performance by up to 2.92% for rare defect classes and 1.48% for common defect classes. It shows that the feature reweighting module actually enhance the presentation of features by embedding information of annotations into the reweighting vector and combining them with features.

Compared with *TL-ff+fr*, our method *TL-SDD* uses distance metric module instead of Fast R-CNN to classify defects which improve the performance by up 6.36% for common defect classes. Figure 9 shows the visualization of the representation of images in metric space in which more visually similar classes tend to have closer representations. For example, the Leak defect is more similar with the Chafed defect than Convex powder defect and the Leak defect is closer to Chafed defect. As shown in Figure 9, the majority of samples are segregated by category, and the model can obtain a better detection performance.

## V. CONCLUSION

In this paper, we propose a novel transfer learning-based method for surface defect detection (TL-SDD). First, we adopt a two-phase training scheme to avoid the overfitting problem caused by training the rare defect classes directly. Second, we propose a novel metric-based surface defect detection model (M-SDD). We design three modules for this model: (1) feature extraction module: combining high-level semantic information with low-level structural information. (2) feature reweighting module: transforming examples to a reweighting vector that indicates the importance of features. (3) distance metric module: learning a metric space in which defects are classified by computing distances to representations of each class. Finally, we validate the performance of the proposed method on a real dataset including surface defects of aluminum profiles. Compared to the baseline methods, defect detection performance was improved by up to 11.98% for rare defect classes. We only take one real data to evaluate our model, but our model can also be applied to other scenarios, such as steel surface, wood surface. In addition, we find that the defect shape is very irregular, and the detection performance of ordinary convolutional neural network on irregular objects may not be very well. In the future, we will try to improve the model by changing convolution and pooling layers.

## ACKNOWLEDGMENT

This work was partially supported by the National Key R&D Program of China(2019YFB1703901), and the National Natural Science Foundation of China (No. 62025205, 62032020, 61725205, 62002292)

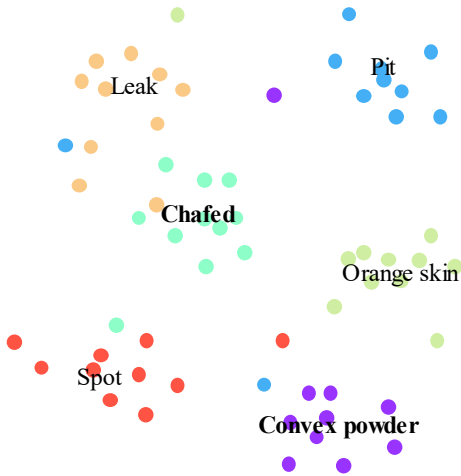

Fig. 9. Visualization of the representation of images in metric space. More visually similar classes tend to have closer representations.